%% file: main.tex
\pdfoutput=1

\documentclass[10pt,twocolumn,letterpaper]{article}

\usepackage[pagenumbers]{cvpr} 

\usepackage{graphicx}
\usepackage{amsmath}
\usepackage{amssymb}
\usepackage{booktabs}
\usepackage{multirow}

\usepackage[pagebackref,breaklinks,colorlinks]{hyperref}

\usepackage[capitalize]{cleveref}
\crefname{section}{Sec.}{Secs.}
\Crefname{section}{Section}{Sections}
\Crefname{table}{Table}{Tables}
\crefname{table}{Tab.}{Tabs.}

\def\cvprPaperID{8017} 
\def\confName{CVPR}
\def\confYear{2023}

\begin{document}

\title{TBP-Former: Learning Temporal Bird's-Eye-View Pyramid for \\ Joint Perception and Prediction in Vision-Centric Autonomous Driving}

\author{Shaoheng Fang$^1$\footnotemark[1] \quad Zi Wang$^{1}$\footnotemark[1] \quad Yiqi Zhong$^2$ \quad Junhao Ge$^{1}$ \quad Siheng Chen$^{1,3}$\footnotemark[2] \quad Yanfeng Wang$^{3,1}$ \\
$^{1}${Cooperative Medianet Innovation Center, Shanghai Jiao Tong University} \\ \quad $^{2}${Department of Computer Science, University of Southern California} \quad $^{3}${Shanghai AI Laboratory} \\ 
$^{1}${\tt\small \{shfang, w4ngz1, cancaries, sihengc, wangyanfeng\}@sjtu.edu.cn}  \quad 
$^{2}${\tt\small \{yiqizhon\}@usc.edu} \\
}




\maketitle

\input{content/0-abstract}
\renewcommand{\thefootnote}{\fnsymbol{footnote}}
\footnotetext[1]{These authors contributed equally to this work.}
\footnotetext[2]{Corresponding author.}
\input{content/1-introduction}

\input{content/2-related_works}

\input{content/3-method}

\input{content/4-experiments}
\input{content/5-conclusion}

\newpage
{\small
\bibliographystyle{ieee_fullname}
\bibliography{egbib}
}



\end{document}

%% file: content/0-abstract.tex
\begin{abstract}

Vision-centric joint perception and prediction (PnP) has become an emerging trend in autonomous driving research. It predicts the future states of the traffic participants in the surrounding environment from raw RGB images. However, it is still a critical challenge to synchronize features obtained at multiple camera views and timestamps due to inevitable geometric distortions and further exploit those spatial-temporal features. To address this issue, we propose a temporal bird’s-eye-view pyramid transformer (TBP-Former) for vision-centric PnP, which includes two novel designs. First, a pose-synchronized BEV encoder is proposed to map raw image inputs with any camera pose at any time to a shared and synchronized BEV space for better spatial-temporal synchronization. Second, a spatial-temporal pyramid transformer is introduced to comprehensively extract multi-scale BEV features and predict future BEV states with the support of spatial priors.  Extensive experiments on nuScenes dataset show that our proposed framework overall outperforms all state-of-the-art vision-based prediction methods. Code is available at:~\href{https://github.com/MediaBrain-SJTU/TBP-Former}{https://github.com/MediaBrain-SJTU/TBP-Former}
\end{abstract}

%% file: content/1-introduction.tex
\vspace{-2mm}
\section{Introduction}
\begin{figure}[t]
    \centering
    \includegraphics[scale=0.32]{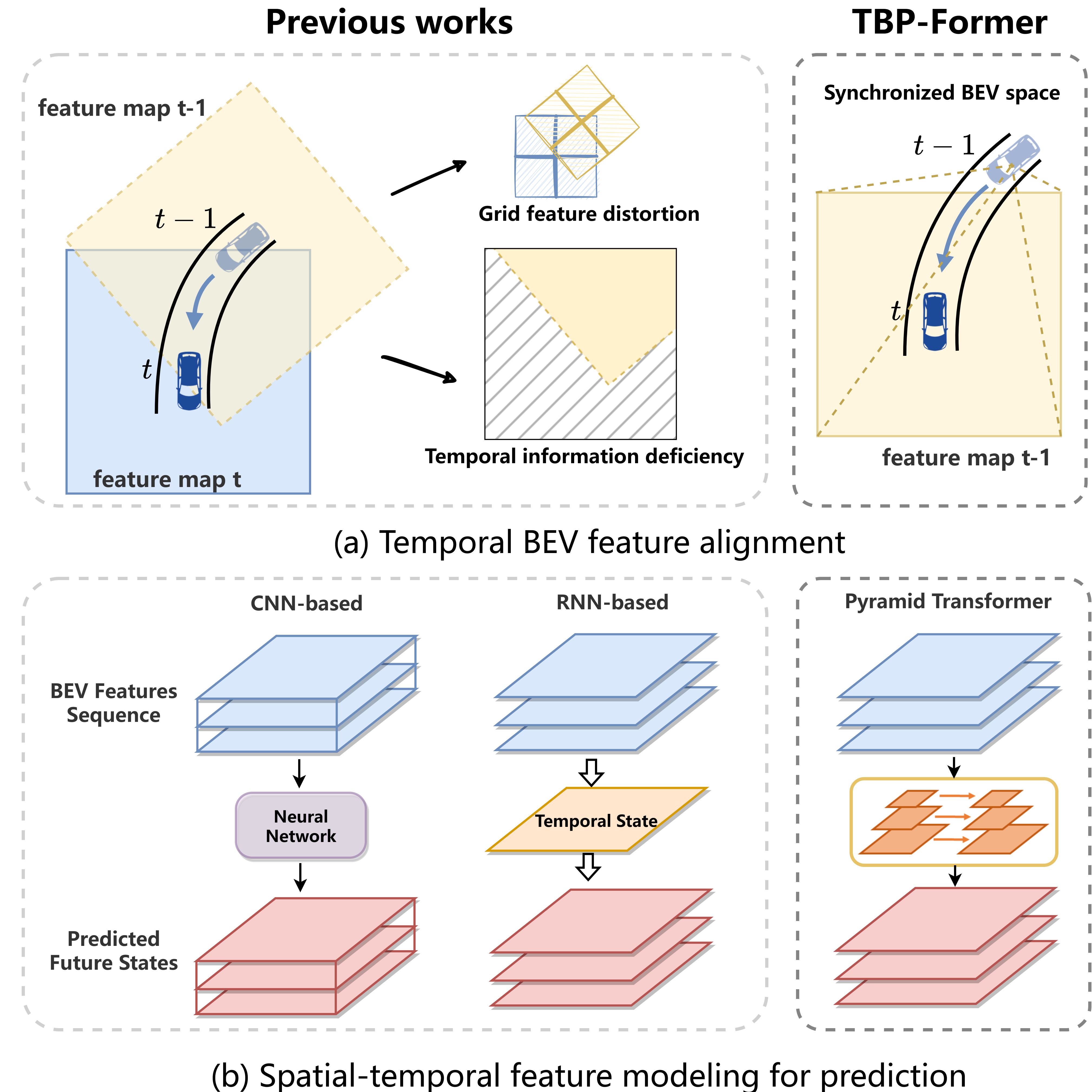}
    \caption{Two major challenges in vision-based perception and prediction are (a) how to avoid distortion and deficiency when aggregating features across time and camera views; and (b) how to achieve spatial-temporal feature learning for prediction. Our Pose-Synchronized BEV Encoder can precisely map the visual features into synchronized BEV space, and Spatial-Temporal Pyramid Transformer extracts feature at multiple scales.}
    \label{fig:warp_problem}
\end{figure} 

As one of the most fascinating engineering projects, autonomous driving has been an aspiration for many researchers and engineers for decades. Although significant progress has been made, it is still an open question in designing a practical solution to achieve the goal of full self-driving. A traditional and common solution consists of a sequential stack of perception, prediction, planning, and control. Despite the idea of divide-and-conquer having achieved tremendous success in developing software systems, a long stack could cause cascading failures in an autonomous system. Recently, there is a trend to combine multiple parts in an autonomous system to be a joint module, cutting down the stack. For example,~\cite{liang2020pnpnet, shah2020liranet} consider joint perception and prediction and~\cite{sadat2020perceive, casas2021mp3} explore joint prediction and planning. This work focuses on joint perception and prediction.

The task of joint perception and prediction (PnP) aims to predict the current and future states of the surrounding environment with the input of multi-frame raw sensor data. The output current and future states would directly serve as the input for motion planning.
Recently, many PnP methods are proposed based on diverse sensor input choices. For example,~\cite{liang2020pnpnet,luo2018faf,casas2018intentnet} take multi-frame LiDAR point clouds as input and achieve encouraging 3D detection and trajectory prediction performances simultaneously. Recently, the rapid development of vision-centric methods offers a new possibility to provide a cheaper and easy-to-deploy solution for PnP. For instance,~\cite{fiery, stretchbev, stp3} only uses RGB images collected by multiple cameras to build PnP systems. Meanwhile, without precise 3D measurements, vision-centric PnP is more technically challenging. Therefore, this work aims to advance this direction.


The core of vision-centric PnP is to learn appropriate spatial-temporal feature representations from temporal image sequences. It is a crux and difficult from three aspects. First, since the input and the output of vision-centric PnP are supported in camera front-view (FV) and bird's-eye-view (BEV) respectively, one has to deal with distortion issues during geometric transformation between two views. Second, when the vehicle is moving, the view of the image input is time-varying and it is thus nontrivial to precisely map visual features across time into a shared and synchronized space. Third, since information in temporal image sequences is sufficiently rich for humans to accurately perceive the environment, we need a powerful learning model to comprehensively exploit spatial-temporal features.

To tackle these issues, previous works on vision-centric PnP consider diverse strategies. For example,~\cite{fiery, beverse} follows the method in~\cite{lift-splat-shoot} to map FV features to BEV features, then synchronizes BEV features across time via rigid transformation, and finally uses a recurrent network to exploit spatial-temporal features. However, due to the image discretization nature and depth estimation uncertainty, simply relying on rigid geometric transformations would cause inevitable distortion; see Fig.~\ref{fig:warp_problem}. Some other work~\cite{wang2021learning} transforms the pseudo feature point cloud to current ego coordinates and then pools the pseudo-lidar to BEV features; however, this approach encounters deficiency due to the limited sensing range in perception. Meanwhile, many works~\cite{fiery, beverse, stp3} simply employ recurrent neural networks to learn the temporal features from multiple BEV representations, which is hard to comprehensively extract spatial-temporal features.

To promote more reliable and comprehensive feature learning across views and time, we propose the temporal bird's-eye-view pyramid transformer (TBP-Former) for vision-centric PnP. The proposed TBP-Former includes two key innovations: i) pose-synchronized BEV encoder, which leverages a pose-aware cross-attention mechanism to directly map a raw image input with any camera pose at any time to the corresponding feature map in a shared and synchronized BEV space; and ii) spatial-temporal pyramid transformer, which leverages a pyramid architecture with Swin-transformer~\cite{liu2021swin} blocks to learn comprehensive spatial-temporal features from sequential BEV maps at multiple scales and predict future BEV states with a set of future queries equipped with spatial priors.

Compared to previous works, the proposed TBP-Former brings benefits from two aspects. First, previous works~\cite{fiery, beverse, stp3, bevformer} consider FV-to-BEV transformation and temporal synchronization as two separate steps, each of which could bring distortion due to discrete depth estimation and rigid transformation; while we merge them into one step and leverage both geometric transformation and attention-based learning ability to achieve spatial-temporal synchronization. Second, previous works~\cite{fiery, wu2020motionnet} use RNNs or 3D convolutions to learn spatial-temporal features; while we leverage a powerful pyramid transformer architecture to comprehensively capture spatial-temporal features, which makes prediction more effective.



To summarize, the main contributions of our work are:
\begin{itemize}
    \item To tackle the distortion issues in mapping temporal image sequences to a synchronized BEV space, we propose a pose-synchronized BEV encoder (PoseSync BEV Encoder) based on cross-view attention mechanism to extract quality temporal BEV features.
    \item We propose a novel Spatial-Temporal Pyramid Transformer (STPT) to extract multi-scale spatial-temporal features from sequential BEV maps and predict future BEV states according to well-elaborated future queries integrated with spatial priors.
    \item Overall, we propose TBP-Former, a vision-based joint perception and prediction framework for autonomous driving.
    TBP-Former achieves state-of-the-art performance on nuScenes~\cite{caesar2020nuscenes} dataset for the vision-based prediction task. Extensive experiments show that both PoseSync BEV Encoder and STPT contribute greatly to the performance. Due to the decoupling property of the framework, both proposed modules can be easily utilized as alternative modules in any vision-based BEV prediction framework.
\end{itemize}

%% file: content/2-related_works.tex
\begin{figure*}[ht]
    \centering
    \includegraphics[scale=0.27]{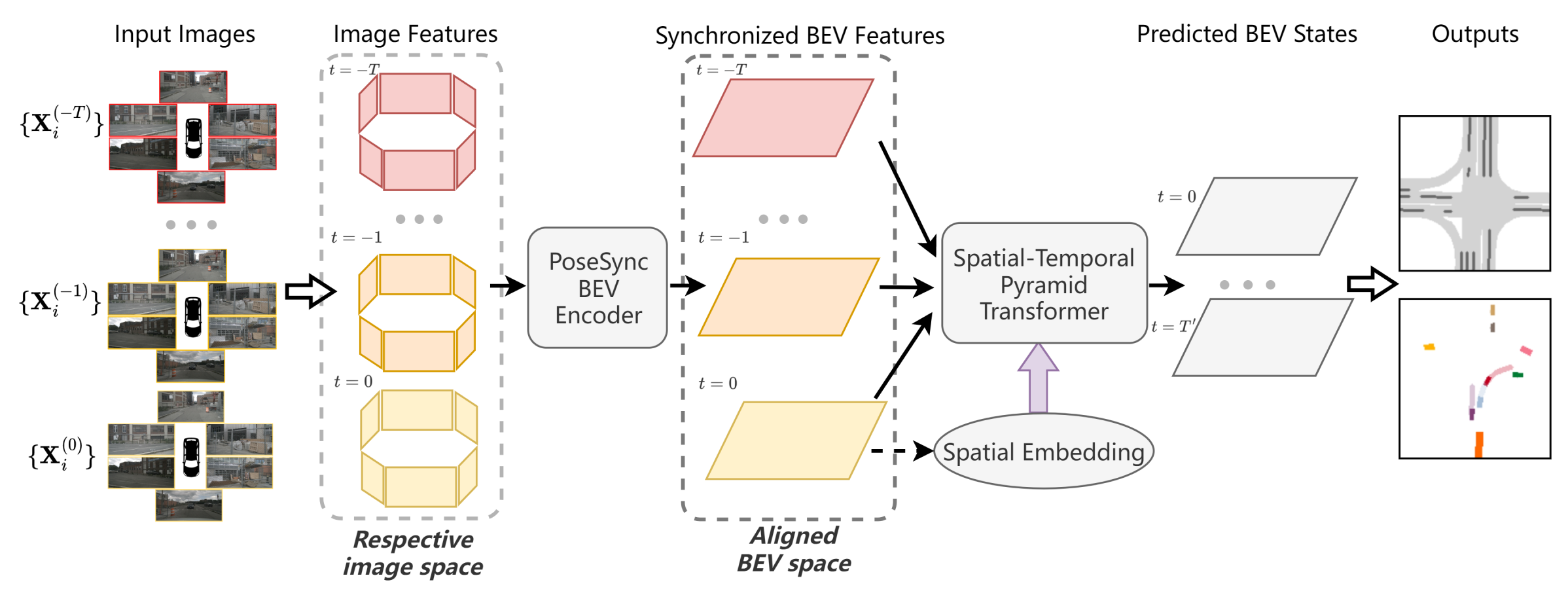}
    \caption{An overview of TBP-Former architecture. Taking consecutive surrounding camera images as inputs, TBP-Former first generates image-space features and uses the PoseSync BEV Encoder to map front-view features to BEV features in a shared and synchronized BEV space. Then the BEV features from multiple frames are processed by the Spatial-Temporal Pyramid Transformer to extract BEV spatial-temporal features and predict future BEV states in order. In this process, high-level scene representations are generated from the last frame BEV feature as spatial priors to guide the prediction. Finally, the well-predicted future states are sent to decoder heads for joint perception and prediction tasks.}
    \label{fig:overview}
     \vspace{-4mm}
\end{figure*} 

\vspace{-2mm}
\section{Related Work}

\subsection{Joint Perception and Prediction}

As the two core system modules of autonomous driving, how to conduct perception and prediction tasks jointly has received a lot of attention. Traditional approaches ~\cite{casas2020spagnn, casas2018intentnet, liang2020pnpnet, luo2018faf, phillips2021deep} formulate this joint task as a trajectory prediction problem that relies on the perception outputs of 3D object detection and tracking. The dependency on intermediate results tends to accumulate errors and lacks the capacity to perceive unknown objects~\cite{wu2020motionnet, wong2020identifying}. 
Subsequently, instance-free methods~\cite{wu2020motionnet, lee2020pillarflow, luo2021pillarmotion, filatov2020any, schreiber2021dynamic, sadat2020perceive} that predict dense future semantic occupancy and flow has become a growing trend to simplify the understanding of dynamic scenes. 
Also, several recent works~\cite{beverse, stp3, fiery} explore joint perception and prediction in the form of dense occupancy and flow using only surrounding camera input. 

In many previous works~\cite{phillips2021deep, sadat2020perceive, casas2018intentnet, liang2020pnpnet}, raster HD (high-definition) maps play an important role as input of the frameworks. HD maps can provide strong priors to guide the predicted results to follow the traffic lanes. However, in practice, HD maps are laborious and costly to produce and require frequent maintenance. Instead of using off-the-peg HD maps, we follow the philosophy of \cite{li2021hdmapnet, casas2021mp3, chen20203d} in predicting online HD maps but propose to learn high-level scene geometry representations from real-time sensor inputs and take these representations as priors for the prediction task.

\subsection{BEV Representations}
BEV representations provide a unified and physical-interpretable way to represent the rich information of road, moving objects and occlusion in a traffic scene, which can be easily utilized for downstream tasks such as motion prediction, planning and control, etc. 
For camera-based methods, how to solve the problem of projecting features from perspective view to BEV is a major challenge. 
Some learnable methods use MLP~\cite{vpn, pon, li2021hdmapnet} or transformer network ~\cite{cvt, bevsegformer} to implicitly reason the relationship between two different views.
LSS~\cite{lift-splat-shoot} proposes the approach of predicting depth distribution per pixel on 2D features, then ‘lifting’ the 2D features according to the corresponding depth distribution to BEV space. Numerous works, aiming at tasks of BEV perception~\cite{bevdet, m2bev, Where2comm:22}, motion prediction~\cite{fiery, stp3, beverse, stretchbev}, lidar-camera fusion~\cite{bevfusion}, etc., follow this form to generate BEV representations.
Also, some methods~\cite{simplebev, bevformer, persformer} explicitly establish the correspondence from BEV location to image-view pixel using homography between image and BEV plane and achieve attractive performance in diverse tasks. 

However, when dealing with temporal information, most methods~\cite{fiery, stp3, beverse, bevformer, stretchbev} warp history BEV representations according to the variation of ego poses. Due to the pre-defined fixed range and size of the BEV grid, rotation and translation operations may cause distortion and out-of-range problems when aligning history BEV maps to current ego coordinates. Though~\cite{qin2022uniformer} introduces a similar operation to us to integrate historical information into the current frame, the design of their model is unable to predict future states. To alleviate these issues, we propose a PoseSync BEV Encoder module based on deformable attention to generate pose-synchronized BEV representations from temporally consecutive image-view input.

\begin{figure*}[ht]
    \centering
    \includegraphics[scale=0.26]{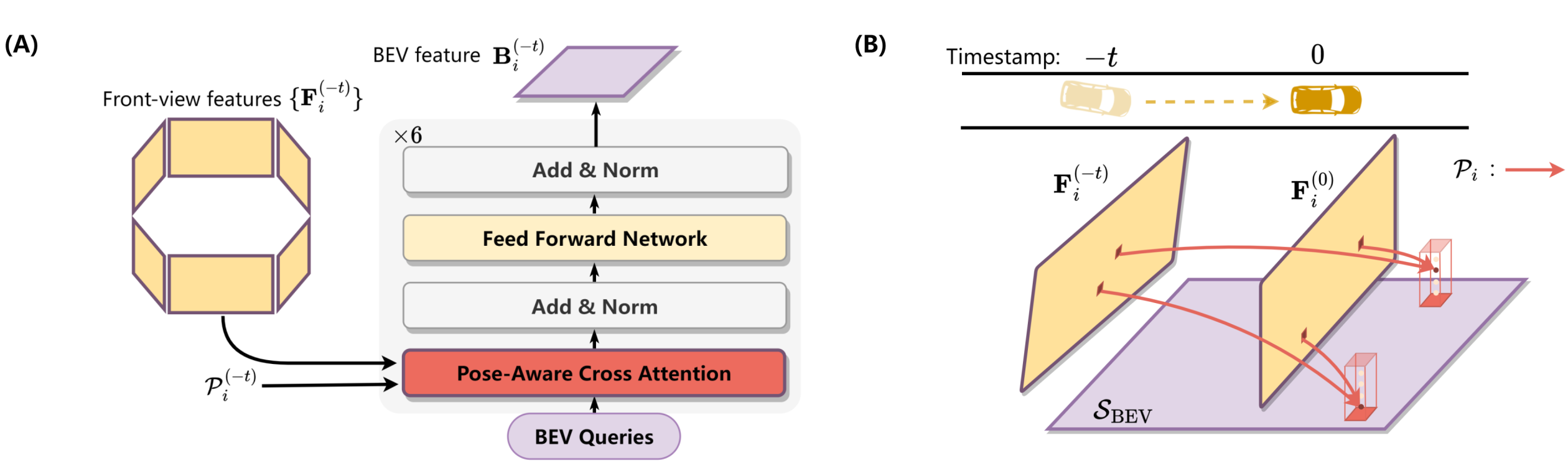}
    \caption{The PoseSync BEV Encoder (A) takes front-view features and camera poses as input and then maps to BEV space. The core to generate BEV features in a synchronized way is the Pose-Aware Cross Attention. Its cross-view attention mechanism is depicted in (B), where front-view features from different frames of a dynamic vehicle are projected into a uniform BEV space.}
    \label{fig:PoseSync BEV Encoder}
     \vspace{-4mm}
\end{figure*} 

\subsection{Spatial-Temporal Modeling}

In the BEV prediction field, how to design a temporal model to aggregate spatial-temporal information is a critical problem. Existing modeling methods can be classified into three categories: RNN-based, CNN-based and transformer-based. RNN-based methods~\cite{fiery, stp3, beverse, stretchbev, sadat2020perceive, CMPNMMP:20} utilize recurrent models such as LSTM~\cite{lstm}, GRU~\cite{gru} to predict the future latent states. Though the recurrent model is powerful to model temporal relationships, it is time-consuming for constraints in the parallelization of computation. Besides, some CNN-based methods~\cite{wu2020motionnet, wang2022sti, luo2018faf, casas2021mp3} concatenate BEV features in the time dimension and take advantage of 3D convolution to extract spatial-temporal features. 

Due to the great power of transformer~\cite{transformer} in sequence modeling, it has shown promise in many temporal modeling tasks such as trajectory prediction~\cite{girgis2021latent, ngiam2021scenetransformer, xu2022GroupNet}, object tracking~\cite{li2022time3d}, video prediction~\cite{gupta2022maskvit, rakhimov2020latent, weissenborn2019scaling}, video interpolation~\cite{lu2022video, geng2022rstt, shi2022video}, etc. For BEV perception, \cite{bevformer, liu2022petrv2} utilize self-attention to model temporal information from multiple frames to boost perception task. \cite{li2022time3d} leverage self-attention to aggregate spatial information and cross-attention to exploit affinities among sequence frames.
To explore the capacity of transformer model in BEV spatial-temporal modeling, we propose a novel Spatial-Temporal Pyramid Transformer (STPT) architecture with future queries for BEV spatial-temporal features extraction and BEV future states prediction.

%% file: content/3-method.tex

\vspace{-2mm}
\section{Methodology}

\subsection{Overview Architecture}
The overall architecture of the proposed TBP-Former is illustrated in Fig.~\ref{fig:overview}. It takes the input of multi-view images with the corresponding camera poses at consecutive $T$ timestamps. The final output includes BEV map segmentation for current scene understanding and occupancy flow for motion prediction. The whole  TBP-Former can be decoupled into three parts: (i) pose-synchronized BEV encoder, which maps raw image sequences into feature maps in a spatial-temporal-synchronized BEV space; (ii) spatial-temporal pyramid transformer, which achieves comprehensive feature learning at multiple spatial and temporal scales; and (iii) a multi-head decoder, which takes the spatial and temporal features to achieve scene understanding and motion prediction. We will elaborate on each part in the following subsections.


\subsection{Pose-Synchronized BEV Encoder}
\label{sec3.2}
Given images collected at multiple time stamps and from various camera poses, we aim to generate the corresponding feature maps in a shared and synchronized BEV space. Different from many previous works that synchronize spatial and temporal information in two separate steps, the proposed pose-synchronized BEV encoder leverage both geometric prior and learning ability to achieve one-step synchronization, alleviating distortion effects. Following the previous transformer-based method~\cite{bevformer}, this encoder adopts a transformer architecture whose core is a novel cross-view attention operation.


\textbf{Front-view feature map.}
Let  $\mathcal{X} = \{ \mathbf{X}_i^{(-t)} \}^{N, T}_{i=1, t= 0}$ be the input multi-frame multi-view images, where $N$ is the number of cameras, $T$ is the number of historical timestamps and $\mathbf{X}_i^{(-t)} \in \mathbb{R}^{ H \times W \times 3}$ is the RBG image captured by the $i$th camera at historical time stamp $t$. Note that each front-view image is associated with a different camera pose. Let $\mathcal{S}_{i}^{(-t)} = \{(u_{i}^{(-t)}, v_{i}^{(-t)})\}_{1,1}^{H,W}$ be the pixel indices of the $i$th camera's front-view space, whose image size is $H \times W$. We feed each RBG image $\mathbf{X}_i^{(-t)}$ into a shared backbone network (our implementation uses ResNet-101~\cite{resnet}) and obtain the corresponding front-view feature map $\mathbf{F}_i^{(-t)} \in \mathbb{R}^{ H^{\prime} \times W^{\prime} \times C}$ with $C$ the channel number, which is also supported on the front-view space $\mathcal{S}_{i}^{(-t)}$.

\textbf{BEV queries.}
Let $\mathcal{S}_{\rm BEV} = \{(x,y)\}_{x=1,y=1}^{X,Y}$ be the BEV grid indices, reflecting the  $X \times Y$ BEV grid space based on the vehicle-ego pose at the current timestamp. Note that $\mathcal{S}_{\rm BEV}$ is the only BEV space we work with in this paper. Let  $\mathbf{Q} \in \mathbb{R}^{ X \times Y \times C}$  be the trainable BEV queries whose element $\mathbf{Q}_{x,y} \in \mathbb{R}^C$ is a $C$-dimensional query feature at the $(x,y)$th geo-location in the BEV space $\mathcal{S}_{\rm BEV}$. We use $\mathbf{Q}$ as the input to query from the front-view feature map $\mathbf{F}_i^{(-t)}$ to produce the corresponding BEV feature map.

\textbf{Cross-view attention.} 
As the key operation in the pose-synchronized BEV encoder, the proposed cross-view attention constructs a feature map in the BEV space $\mathcal{S}_{\rm BEV}$ by absorbing information from the corresponding pixels in the front-view feature map.

Let $\mathcal{P}_i^{(-t)}: \mathcal{S}_{\rm BEV} \times \mathcal{Z} \rightarrow  \mathcal{S}_{i}^{(-t)}$ be a project operation that maps a BEV index with a specific height index to a pixel index in the $i$th camera's front view at historical timestamp $t$; that is,
\begin{equation*}
    (u_{i}^{(-t)}, v_{i}^{(-t)} ) = \mathcal{P}_i^{(-t)} \Big( (x,y,z) \Big),
\end{equation*}
where $z \in \mathcal{Z} = \{1,\cdots, Z\}$. The project operation $\mathcal{P}_i^{(-t)}$ builds the geometric relationship between the BEV and a front view. The implementation of $\mathcal{P}_i^{(-t)}$ works as 
\begin{equation*}
\begin{aligned}
z^{(-t)}_{i} \cdot \begin{bmatrix}
    u^{(-t)}_{i} \\[4pt]
    v^{(-t)}_{i} \\[4pt]
    1
\end{bmatrix} =  \mathcal{T}_{\mathcal{S}_{\rm BEV} \rightarrow  \mathcal{S}_{i}^{(-t)}} \cdot
\begin{bmatrix}
    x \\
    y \\
    z \\
    1
\end{bmatrix},
\end{aligned}
\label{eq:projection}
\end{equation*}
where $\mathcal{T}_{\mathcal{S}_{\rm BEV} \rightarrow  \mathcal{S}_{i}^{(-t)}} \in \mathbb{R}^{3 \times 4}$ is a transformation matrix that can be calculated by camera's intrinsic/ extrinsic parameters and vehicle-ego pose.

Based on the project operation, the cross-view attention can trace visual features in the front view through a BEV index. Let $\mathbf{B}_i^{(-t)} \in \mathbb{R}^{ X \times Y \times C}$ be the BEV feature map associated with the RBG image $\mathbf{X}_i^{(-t)}$. The $(x,y)$th element of the BEV feature map is obtained as
\begin{equation}
\label{eq:cross_view_attention}
(\mathbf{B}_i^{(-t)})_{x,y} \ = \ \sum_{z} f_{\rm DA} \left( \mathbf{Q}_{x,y}, \mathcal{P}_i^{(-t)}(x,y,z), \mathbf{F}_i^{(-t)} \right),
\end{equation}
where, $f_{\rm DA}(\cdot)$ represents the deformable attention operation~\cite{deformabledetr}. It allows BEV query $Q_{x,y}$ only to interact with the front-view feature $\mathbf{F}_i^{(-t)}$ within its regions of interest, which is sampled around the reference point calculated by $\mathcal{P}_i^{(-t)}$. 
Since one BEV index might lead to multiple pixel indices in the front-view image because of various height possibilities in the 3D space. We thus sum over all possible heights in~\eqref{eq:cross_view_attention}.  To further aggregate BEV feature maps across all the $N$ camera views, we simply take the average; that is, the BEV feature map at historical timestamp $t$ is $\mathbf{B}^{(-t)} = \frac{1}{N} \sum_i \mathbf{B}_i^{(-t)}$.
Note that all the front-view features across time and from multiple cameras are synchronized into the same BEV space in one step~\eqref{eq:cross_view_attention}, leading to less information distortion or deficiency issues.

We can successively apply the cross-view attention followed by feed forward networks and normalization layers for multiple times. Finally, we order BEV feature maps at multiple timestamps and obtain a temporal BEV feature map $\mathcal{B} = [ \mathbf{B}^{(0)}, \mathbf{B}^{(-1)}, ..., \mathbf{B}^{(-T)} ] \in \mathbb{R}^{(T+1) \times X \times Y \times C} $.

\subsection{Spatial-Temporal Pyramid Transformer}
\label{sec3.3}

\begin{table*}[ht]
\centering
\begin{tabular}{c|c|cc|cc|c}
\toprule
\multirow{2}{*}{Method}                       & RGB            & \multicolumn{2}{c|}{Future semantic seg.}                       & \multicolumn{2}{c|}{Future instance seg.}                       & \multirow{2}{*}{~~~FPS~~~} \\
& Resolution     & IoU (Short)& IoU (Long)  & VPQ (Short)             & VPQ (Long) &     \\ 
\hline
\hline
FIERY\cite{fiery}            & 224$\times$480 & 59.4                           & 36.7                           & 50.2                           & 29.9                           & 1.56                                 \\
StretchBEV \cite{stretchbev} & 224$\times$480 & 55.5                           & 37.1                           & 46.0                           & 29.0                           & 1.56                                 \\
ST-P3\cite{stp3}             & 224$\times$480 & -                              & 38.9                           & -                              & 32.1                           & 1.43                                 \\
BEVerse \cite{beverse}       & 256$\times$704 & 60.3                           & 38.7                           & 52.2                           & 33.3                           & 1.96                                 \\ \midrule
TBP-Former                                          & 224$\times$480 & \textbf{64.7} & \textbf{41.9} & \textbf{56.7} & \textbf{36.9} & \textbf{2.44}                                 \\ \bottomrule
\end{tabular}
\caption{\textbf{Prediction results on nuScenes~\cite{caesar2020nuscenes} validation set.} Intersection-over-Union (IoU) is used for future semantic segmentation and Video Panoptic Quality (VPQ) for future instance segmentation. Results are reported under two settings: short ($30m \times 30m$) range and long ($100m \times 100m$) range. Frame Per Second (FPS) means the inverse of inference time. All methods are tested under the same settings on a single NVIDIA A100. Our TBP-Former achieves SOTA performance and is still more computationally efficient than other methods.}
\label{tab:prediction_result}
 \vspace{-3mm}
\end{table*}

\begin{figure}[t]
    \centering
    \includegraphics[scale=0.38]{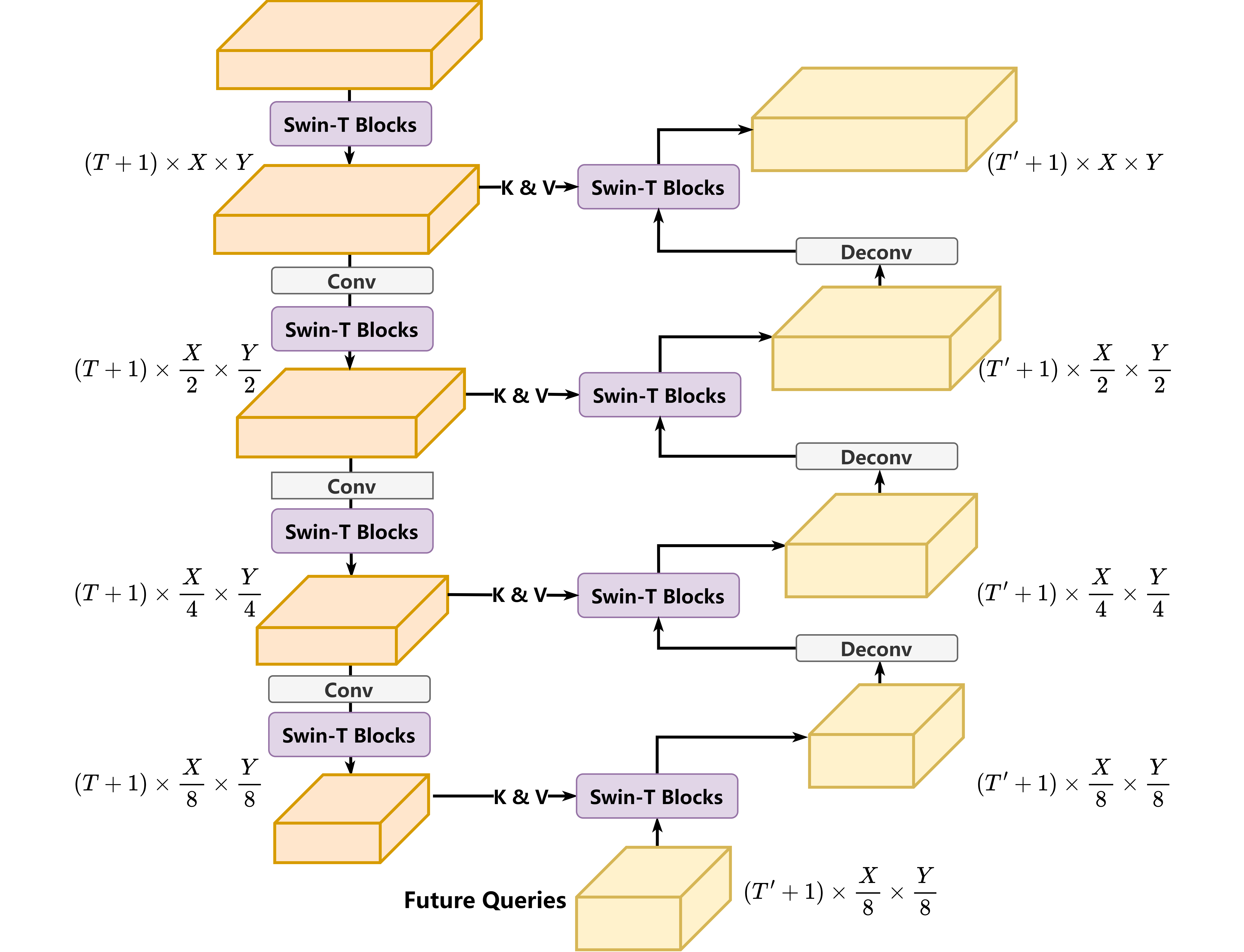}
    \caption{The network architecture of Spatial-Temporal Pyramid Transformer (STPT). Each encoder layer consists of an optional convolutional block for downsampling and Swin Transformer Blocks, while each decoder layer contains Swin Transformer Blocks and a deconvolutional block for upsampling. In the decoding process, we pre-define a set of future queries to represent future BEV states and query the features from encoders. }
    \label{fig:temporal_model}
     \vspace{-2mm}
\end{figure} 

We further propose a novel spatial-temporal pyramid transformer (STPT) to learn spatial-temporal features more comprehensively and produce the future BEV states. The detailed structure of STPT is depicted in Fig.~\ref{fig:temporal_model}.

\textbf{Temporal BEV pyramid feature learning.} We encode the input temporal BEV feature map $\mathcal{B}$ with four hierarchical layers. Each encoder layer is composed of an optional convolution layer with stride 2 to downsample the features and a swin transformer encoder, which is a stack of Swin Transformer blocks~\cite{liu2021swin}. In our implementation, the window size of Swin-T blocks is set as $(4, 4)$. We can then obtain multi-scale spatial-temporal features $\mathcal{B}_s \in \mathbb{R}^{(T+1) \times \frac{X}{2^s} \times \frac{Y}{2^s}  \times C}, ~s=0,1,2,3$.

\textbf{Future BEV queries.}
Future BEV queries are defined to represent future BEV states and query the generated multi-scale spatial-temporal features. There is a set of learnable future queries $\{\mathbf{Q}^{(t)}\}, ~t=0, \dots, T^{\prime}$, where $\mathbf{Q}^{(t)}$ has the same spatial dimension $\frac{X}{8} \times \frac{Y}{8}$ as $\mathcal{B}_3$.
Separate learning embeddings are employed for future queries to differentiate the predicted BEV states over time. Additionally, a map feature generator is applied to generate high-dimensional features from $\mathbf{B}^{(T)}$ in order to extract information about the scene's geometry. To be specific, the same structure and parameters of the hdmap decoder head (excluding the last linear layer) are reused. The resulting map feature is added to all future queries to provide spatial information priors.

\textbf{Future BEV state prediction.} The decoding process contains corresponding four hierarchical layers as the encoding process. Unlike the encoding process, $\{\mathbf{Q}^{(t)}\}$ is used as the query input of Swin-T block and performs cross attention with the encoded features $\mathbf{E}_3$. After the first decoding layer, the output of each layer is used as the query input of the next layer. Similar to encoding layers, the deconvolution layer is optionally applied to upsample the decoded features. The simplified process can be written as 
\vspace{-4mm}

$$ \mathcal{D}_s = \left\{ 
\begin{array}{rcl}
\begin{aligned}
 &\mathrm{SwinT}(\mathcal{B}_3, \{\mathbf{Q}^{(t)}\}), & s =3   \\
 & \mathrm{SwinT}(\mathcal{B}_s, \mathrm{DeConv}(\mathbf{D}_{s+1})), & s =0, 1, 2 \\
\end{aligned}
\end{array} \right. $$
\vspace{-1mm}
where $\mathcal{D}_s  \in  \mathbb{R}^{(T'+1)  \times \frac{X}{2^s} \times \frac{Y}{2^s} \times C}, ~s = 0,1,2,3$ are decoded features.
The future temporal BEV feature map at the $0$th scale is a temporal sequence of final predicted BEV states; that is, $\mathcal{D}_0 =  [\mathbf{B}^{(0)}_*,\mathbf{B}^{(1)}_*,..., \mathbf{B}^{(T')}_* ]$, where $\mathbf{B}^{(t)}_* \in \mathbb{R}^{X \times Y \times C}$ is the BEV state at future time stamp $t$.

\begin{table}[t]
\centering
\begin{tabular}{c|c|cc}
\toprule
Method    & Temp. & Veh. IoU                      & Ped. IoU \\
\hline
\hline
VED    \cite{ved}                      &        &  23.3 & 11.9      \\
VPN  \cite{vpn}                        &         & 28.2 & 10.3      \\
PON    \cite{pon}                      &         &  27.9 & 13.9      \\
LSS    \cite{lift-splat-shoot}                      &         &  34.6 & 15.0      \\
CVT  \cite{cvt}                        &         & 36.0                         & -          \\
Image2Map \cite{saha2022translating}  &         & 40.2                         & -          \\
BEVFormer \cite{bevformer}                    &         & 44.4                         & -          \\
IVMP    \cite{wang2021learning}                     & \checkmark        & 36.8                        & 17.4      \\
FIERY  \cite{fiery}         & \checkmark        & 38.2                         & 17.2      \\
ST-P3      \cite{stp3}                  & \checkmark        & 40.1                        & 14.5      \\
\midrule
TBP-Former \textit{static}            &         & 44.8                        & 17.2      \\
TBP-Former    & \checkmark        & \textbf{46.2}                        & \textbf{18.6}     \\
\bottomrule
\end{tabular}
\caption{\textbf{Perception results on nuScenes~\cite{caesar2020nuscenes} validation set.} Results of vehicles and pedestrians are compared by segmentation IoU. Temp. indicates whether temporal information is involved.}
\label{tab:perception_result}
 \vspace{-5mm}
\end{table}

\subsection{Multi-head decoder}
\label{sec3.4}

The future temporal BEV feature map is fed into the multi-task decoder heads to generate various outputs for dynamic scene understanding, see Fig.~\ref{fig:result}. We follow the output setting in~\cite{fiery} that predicts BEV semantic segmentation, instance center, instance offset, and future flow for joint perception and prediction. Meanwhile, we set up an additional HD map decoder head to predict basic traffic scene elements including drivable areas and lanes. The map decoder head can not only provide scene information for subsequent planning and control modules but also give guidance to the prediction process, see sec.~\ref{sec3.3}.

%% file: content/4-experiments.tex
\vspace{-3mm}
\section{Experiments}

\begin{table*}[ht]
\centering
\begin{tabular}{c|cccc|cc|cc}
\toprule
 &  &  & &  & \multicolumn{2}{c|}{Future semantic seg. }  & \multicolumn{2}{c}{Future instance seg.}  \\
Exp. & Warp. & Sync. & SLQ & SPE &Short (IoU) &  Long (IoU) &  Short (VPQ) &  Long (VPQ) \\
\hline
\hline
1   & \checkmark                 &                 &                                                             &                                                                  & 58.7              & 38.4             & 50.6              & 31.8             \\
2   & \checkmark                 &                 & \checkmark                                                           &                                                                  & 60.8              & 38.6             & 52.4              & 33.4             \\
3   &\checkmark                 &                 & \checkmark                                                           & \checkmark                                                                & 62.0              & 40.7             & 53.2              & 34.3             \\
\hline
\hline
4   &                   &\checkmark               &                                                             &                                                                  & 63.0              & 40.8             & 54.1              & 34.3              \\
5   &                   & \checkmark               & \checkmark                                                          &                                                                  & 63.8              & 41.1              & 55.8              & 35.8              \\
6   &                   &\checkmark               & \checkmark                                                           & \checkmark                                                               & \textbf{64.7}              & \textbf{41.9}             & \textbf{56.7}              & \textbf{36.9}             \\
\bottomrule
\end{tabular}
\caption{\textbf{Ablation of our proposed architecture.} Ablation results for our PoseSync BEV Encoder (Sync.), the learnable future queries, and the spatial embedding are presented. Exp. 1-3 use the traditional warping methods to align temporal BEV features. Separate learnable queries (SLQ) represent using separate learnable future queries instead of utilizing the same query with temporal positional encoding. Spatial positional embedding (SPE) represents using spatial scene representations in future prediction queries.}
\vspace{-4mm}
\label{tab:module_ablation}
\end{table*}

\begin{table}[ht]
\centering
\begin{tabular}{c|c|ccc}
\toprule
Temporal model                                                   & IoU                            & VPQ                            & VRQ                            & VSQ                            \\ \hline
\hline
$\text{MotionNet}^{\dag}$~\cite{wu2020motionnet}   & 35.4                           & 30.6                           & 43.1                           & 71.1                           \\
$\text{FIERY}^{\dag}$~\cite{fiery}               & 38.3                           & 32.1                           & 45.4                           & 70.7                           \\
$\text{BEVerse}^{\dag}$~\cite{beverse}           & 40.2                           & 34.0                           & 48.0                           & 70.9                           \\ \hline
TBP-Former                                                             & \textbf{41.9} & \textbf{36.9} & \textbf{51.5} & \textbf{72.6} \\ \hline
\end{tabular}
\caption{\textbf{Ablation for the prediction model.} $\dag$: We use MotionNet, FIERY and BEVerse to replace our prediction model for comparison, and the BEV encoder and task heads are the same. Besides IoU and VPQ, we also use Video Recognition Quality (VRQ) and Video Segmentation quality (VSQ) for evaluation. }
\label{tab:temporal_ablation}
\vspace{-1mm}
\end{table}

\begin{table}[ht]
\centering
\begin{tabular}{cc|cc|cc}
\toprule
\multicolumn{2}{c|}{Augmentation} & \multicolumn{2}{c|}{Perception}                   & \multicolumn{2}{c}{Prediction}           \\ 
  Cam  &  BEV       & \multicolumn{1}{c}{Veh.} & Ped. & \multicolumn{1}{c}{IoU} & VPQ\\ 
\hline
\hline
&      & \multicolumn{1}{c}{45.0}  & 17.7 & \multicolumn{1}{c}{40.5}     & 34.4     \\
\checkmark  &   & \multicolumn{1}{c}{44.8} & 18.5           & \multicolumn{1}{c}{40.9}     & 35.3     \\
& \checkmark   & \multicolumn{1}{c}{45.3} & 18.6 & \multicolumn{1}{c}{41.4}     & 35.6     \\
\checkmark  & \checkmark & \multicolumn{1}{c}{\textbf{46.2}}  & \textbf{18.6}       & \multicolumn{1}{c}{\textbf{41.9}}  & \textbf{36.9} \\ \bottomrule
\end{tabular}
\caption{\textbf{Ablation for data augmentation strategies.} Perception of Vehicles and Pedestrians with different data augmentation strategies are evaluated on segmentation IoU. Prediction results are evaluated on segmentation IoU and Video Panoptic Quality. }
\label{tab:aug_ablation}
    \vspace{-4mm}
\end{table}

\subsection{Dataset and settings}
We use nuScenes~\cite{caesar2020nuscenes} datasets to evaluate our approach. NuScenes contains 1000 scenes, each of which has 20 seconds annotated at 2Hz. In nuScenes, the images are captured by 6 cameras with a small overlap in the field of view, which guarantees the cameras cover the full 360° field of view. For model input, raw camera images with the size of $900 \times 1600$ are resized and cropped to a resolution of $224 \times 480$. We follow the training and evaluating settings used in previous methods~\cite{fiery, stp3, stretchbev, beverse} for fair comparisons, which use 1.0 second past states and current state to predict 2.0 seconds of the future states. It corresponds to predicting 4 future frames based on 3 observed frames. The 
size of the generated BEV grid map is $200 \times 200$. Each grid has a range of $0.5m \times 0.5m$, which means the perception and prediction range is $100m \times 100m$.

For training, we use AdamW~\cite{loshchilov2017decoupled} with a weight decay $0.01$ to optimize the models. The learning rate is initialized as $10^{-4}$ and decays with a cosine annealing scheduler~\cite{loshchilov2016sgdr}. All models are trained on 4 NVIDIA A100 GPUs for $10$ epochs. 

\subsection{Metrics}
Following previous works~\cite{fiery, beverse, stp3, stretchbev}, we mainly use two metrics for evaluation. 
The first is Intersection over Union (IoU), which measures the quality of segmentation at each frame. The second is Video Panoptic Quality (VPQ), which is used to measure the consistency of the detected instances over time and the accuracy of the segmentation. 
The formula is shown below:
\begin{equation*}
 \vspace{-3mm}
\begin{aligned}
\textrm{VPQ}=\sum^{H}_{t=0} \frac{\sum_{(p_t, q_t) \in TP_t}\textrm{IoU}(p_t, q_t)}{\left| TP_t \right| + \frac{1}{2} \left| FP_t \right| + \frac{1}{2} \left| FN_t \right|}
\end{aligned}
\end{equation*}
where $H$ is the sequence length, $TP_t$ represents the set of true positives, $FP_t$ represents the set of false positives and $FN_t$ represents the set of false negtives at timestamp $t$.

\vspace{-1mm}
\subsection{PnP results}
\vspace{-1mm}

\textbf{Perception and Prediction.}
Table~\ref{tab:prediction_result} compares TBP-Former with other methods of perception and prediction task based on multi-view cameras. We see that \textrm{i}) we achieve state-of-the-art performance and exceed previous methods by a large margin. \textrm{ii}) Even though BEVerse has larger RGB resolutions, TBP-Former still surpasses their performance on IoU by \textbf{7.3\%/8.3\%} for short/long settings, respectively. TBP-Former also improves the VPQ by \textbf{12.1\%/10.8\%}. \textrm{iii}) Apart from the performance improvement, TBF-Former also has a larger FPS compared to other methods. Its inference speed is 25\% faster than BEVerse's.

\begin{figure*}[t]
    \centering
    \includegraphics[scale=0.60]{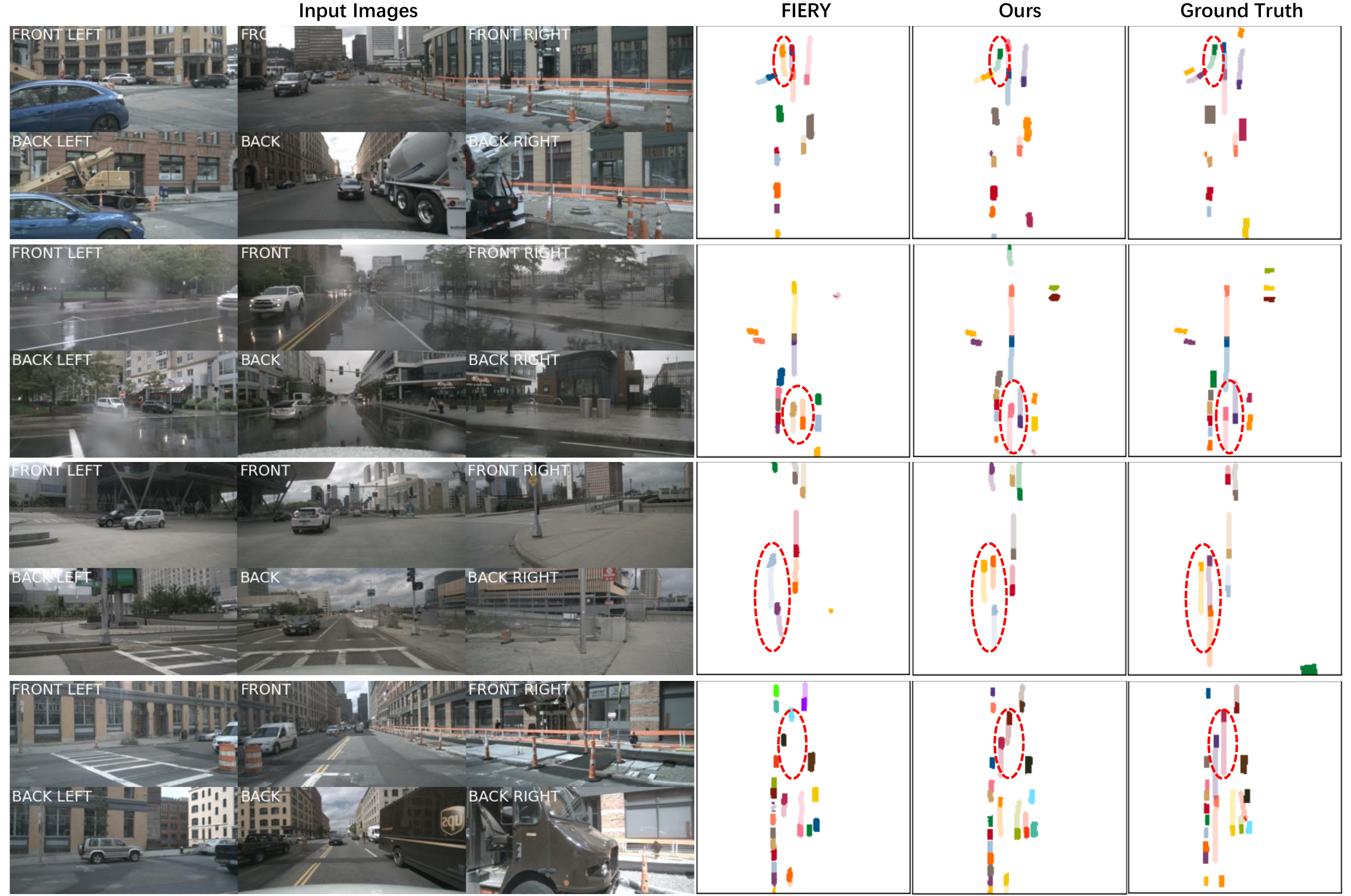}
    \caption{Demonstration of our results compared with FIERY and Ground Truth. 
    Different vehicles are assigned with different colors in order to make a distinction. The darker parts represent the perception of the current frame, and the lighter parts represent the prediction of the vehicles in future frames. The visualization is based on the predicted occupancy and flow. }
    \label{fig:result}
    \vspace{-3mm}
\end{figure*}

Fig.~\ref{fig:result} shows the visualization results of our proposed method. We see that \textrm{i}) almost all the objects are detected correctly except for those occluded ones. \textrm{ii}) TBP-Former is capable of capturing the motion information in past frames and precisely predicting the vehicles' trajectories by occupancy and flow. Compared with FIERY~\cite{fiery}, TBP-Former is closer to the ground truth. \textrm{iii}) TBP-Former does a better job than FIERY when predicting vehicles' turning.

\textbf{Perception Only.} 
Table~\ref{tab:perception_result} compares the results of plenty of state-of-the-art methods on perception (segmentation) task.  We see that our static model, which does not contain temporal information, can achieve 44.8 and 17.2 IoU of vehicles and pedestrians. With the input of temporal sequences, the performance improves further since auxiliary information is provided for better perception. The state-of-the-art results prove the effectiveness of the novel design of our Pose-synchronized BEV encoder.

\vspace{-2mm}
\subsection{Ablation}
\vspace{-1mm}

\textbf{Effectiveness of PoseSync View Projection.} 
The Exp.~1\&4,~2\&5,~3\&6 in Table~\ref{tab:module_ablation} compare the proposed PoseSync View Projection and the existing feature warping methods. We see that the proposed method always achieves better performance when other settings remain the same. The reasons are that: \textrm{i}) PoseSync View Projection based on Deformable Attention can guarantee the precise correspondence between BEV grids and image features. \textrm{ii}) Our projection method can alleviate distortion and our-of-range issues when synchronizing sequential BEV features. 


\textbf{Effectiveness of the designed future queries.} 
In Exp.~1\&3 in Table~\ref{tab:module_ablation}, we utilize the identical query with temporal positional encoding for future queries. Exp.~2\&4 in Table~\ref{tab:module_ablation} demonstrate that using separate learnable embedding for future queries can achieve better performance.
Exp.~3\&6 in Table~\ref{tab:module_ablation} validate the efficacy of the proposed spatial priors for future queries. The generated high-dimensional map features provide the prediction model with useful geographic information. The additional spatial information can aid the prediction and lead to better scene forecasting.

\textbf{Effectiveness of STPT.} Table~\ref{tab:temporal_ablation} compares STPT with popular CNN-based~\cite{wu2020motionnet} and RNN-based~\cite{fiery, beverse} methods. We keep all the settings the same except for temporal modeling. To be specific, the size of input images ($224 \times 480$), image backbones and BEV feature extractor are the same. And then we plug their temporal models into our architecture. We see that \textrm{i}) STPT model performs better in all four metrics, including semantic segmentation IoU and three instance segmentation metrics from the video prediction area.  \textrm{ii}) Our reproduced temporal models achieve higher performance than the original implements. This further validates the effectiveness and power of our BEV feature extractor.

\textbf{Data Augmentation.} We perform both image-view and BEV augmentations. The image-view augmentations include random scaling, rotation and flip of the input images. The BEV augmentations include similar operations on both BEV representations and corresponding ground truth labels. Table~\ref{tab:aug_ablation} compares the results of different data augmentation strategies. We see that \textrm{i}) both augmentation methods improve the performance when used separately. \textrm{ii}) The combination of two methods works better than any single approach. Introducing data augmentation strategies is beneficial to the model's robustness and generalization ability.

%% file: content/5-conclusion.tex
\vspace{-2mm}
\section{Conclusion}
\vspace{-2mm}
This paper proposes a novel TBP-Former for vision-centric joint perception and prediction. We design a pose-synchronized BEV encoder module using a cross-view attention mechanism to solve the distortion issues in previous works. Furthermore, we propose a powerful spatial-temporal pyramid transformer for BEV feature extraction and BEV state prediction. Experiments show that i) TBP-Former improves the prediction performance over state-of-the-art methods significantly; and ii) both PoseSync BEV Encoder and STPT contribute to better performances. 

\textbf{Acknowledgement}  This research is partially supported by National Natural Science Foundation of China under Grant 62171276 and the Science and Technology Commission of Shanghai Municipal under Grant 21511100900 and 22DZ2229005.